\documentclass{article}
\usepackage{spconf,amsmath,graphicx}
\usepackage{comment}
\usepackage{algorithm,algpseudocode}

\title{Convolutional Mixture Density Recurrent Neural Network for Predicting User Location with WiFi Fingerprints}
\name{Weizhu Qian, Fabrice Lauri, Franck Gechter\thanks{Email: weizhu.qian@utbm.fr}}
\address{CIAD, Univ. Bourgogne Franche-Comt\'e \\
UTBM, F-90010, Belfort, France \\ }

\begin{document}
%
\maketitle
\begin{abstract}

Predicting smartphone users activity using WiFi fingerprints has been a popular approach for indoor positioning in recent years. However, such a high dimensional time-series prediction problem can be very tricky to solve. To address this issue, we propose a novel deep learning model, the convolutional mixture density recurrent neural network (CMDRNN), which combines the strengths of convolutional neural networks, recurrent neural networks and mixture density networks. In our model, the CNN sub-model is employed to detect the feature of the high dimensional input, the RNN sub-model is utilized to capture the time dependency and the MDN sub-model is for predicting the final output. For validation, we conduct the experiments on the real-world dataset and the obtained results illustrate the effectiveness of our method.

\end{abstract}

\section{Introduction}
\label{sec: Intro}

Location based service (LBS) has a significant meaning for applications like location-based advertising, outdoor/indoor navigation and social networking, etc. With the help of the advancement of the smartphone technology in recent decades, smartphone devices are integrated with various built-in sensors, such as GPS modules, WiFi modules, cellular modules, etc. By acquiring the data from these sensors, researchers are capable of studying human activities. Among these research topics, user location analysis and prediction has been a research foci. There are several possible classes of methods can be applied for such subjects. Since the GPS equipment can provide relatively accurate outdoor position information, GPS-based methods are favored by many researchers \cite{cho2016exploiting}, \cite{yu2017modeling}. However, such methods are not suitable for indoor positioning. A more applicable approach is to make use of WiFi fingerprints of the smartphone devices. In this case, received signal strength indicator (RSSI) of WiFi access points scanned by the mobile phones is adopted to identify the locations of the users.    

In literature, researchers have exploited various kinds of machine learning techniques, both conventional learning methods and deep learning, on location recognition and prediction with WiFi fingerprints. In the previous work of \cite{bozkurt2015comparative}, \cite{cramariuc2016clustering}, \cite{ferris2007wifi}, \cite{hahnel2006gaussian}, \cite{yiu2015gaussian}, the researchers adopted several conventional machine learning methods for classification, clustering and regression tasks, for instances, decision trees, K-nearest neighbors, naive Bayes, neural networks, K-means, the affinity clustering algorithm, Gausian Process, etc. Deep learning based methods, such as convolutional neural networks ), autoencoders and recurrent neural networks also have been applied in WiFi based positioning methods. Note that in real world, a building may be equipped with a relatively large number of WiFi hotpots to provide good wireless connections. Consequently, this leads to the issue of high dimensionlity. Naturally, some deep-learning based dimension-reduction methods like auto-encoders can be used before the classification or prediction tasks \cite{nowicki2017low}, \cite{song2019novel}, \cite{kim2018scalable}.

 In our work, we attempt to utilize the WiFi fingerprints to predict the accurate user location. This task can be regarded as a high dimensional time-series prediction. The training inputs of our model are the RSSI value vectors and the training targets are the future coordinate values (2D). Though, some previous researchers have used CNNs or RNNs for accurate indoor localization \cite{ibrahim2018cnn}, \cite{song2019novel}, \cite{hoang2019recurrent}, we argue that the models with ordinary euclidean-distance loss functions (for instance, mean square errors) are not capable of overcoming the serve nonlinearity of the data caused by the signal-fading and multi-path effects, WiFi signals are not always stable \cite{hoang2019recurrent}.

In contrast with the aforementioned methods, we devise an innovative hybrid deep learning structure, the convolutional mixture density recurrent neural network (CMDRNN). Compared to other existing models, first, our approach does not need to pre-train an autoencoder to reduce the dimension, instead, we deploy a CNN structure to detect the feature of input. Second, we make use of a RNN to exhibit the temporal dynamics behavior of user trajectory. As for the final output of the network, we employ a MDN structure to calculate the conditional probability density rather than predict the output directly as other conventional neural networks. Therefore, our model consists of three sub-models, a CNN sub-structure, a RNN sub-structure and a MDN sub-structure, which enables our model to tackle complicated time-series prediction problem as the WiFi fingerprints based location prediction. The main contributions of our work are summarized as follows. $1)$ In order to predict user location with WiFi fingerprints, we devise a novel hybrid deep-learning model, in which the advantages of CNNs, RNNs and MDNs are merged. $2)$ We conduct the evaluation experiments on a real-world dataset to test our model and compare other models as well. The final results show the superiority of our method.  

The reminder of the paper is organized as follows. In Section ~\ref{sec: Method}, the proposed model is introduced. In Section ~\ref{sec: Experiments} presents the validation experiments and the results with the real user data. Finally, we draw the conclusions and discuss about the possible future work in Section ~\ref{sec: Conclusions}.

\section{Proposed Method}
\label{sec: Method}

\subsection{Proposed Model Overview}

\begin{figure*}[!t]
\centering
\includegraphics[width= 0.9\linewidth]{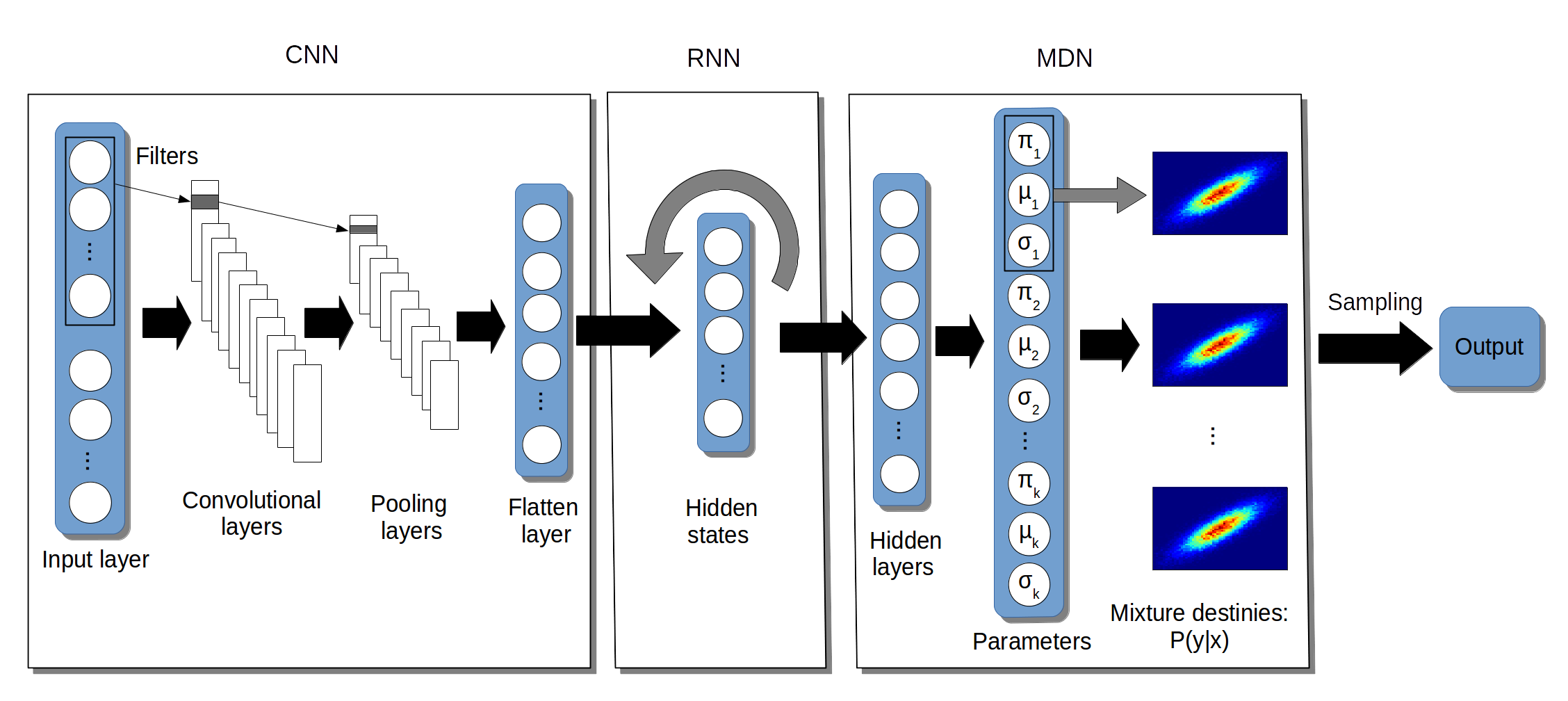}
\caption{Convolutional Mixture Density Recurrent Neural Network.}\label{Fig:CMDRNN}
\end{figure*}

Combined with the merits of three different deep neural networks, we devise a novel deep neural network architecture, which is called the convolutional mixture density recurrent neural network (CMDRNN). The proposed model is composed of an one-dimensional convolutional neural network, a recurrent neural network and a mixture density network. The whole structure of our model is demonstrated in Fig \ref{Fig:CMDRNN}.

\subsection{1D Convolutional Neural Network Sub-structure}

The first step of our approach is to capture the feature of the high dimensional input. In practice, we find that, for each vector, only a few elements has real meaningful values, whereas most elements are not activated. It is easy to understand because the smartphone only can detect very limited number of WiFi access points at each time in a building with a considerable number of WiFi access points. As a consequence, the feature of the input is hard to be captured. To deal with this issue, we resort to the powerful deep-learning technique, convolutional neural networks (CNNs) \cite{lecun1998gradient}. CNNs are widely used for tasks such as imagine processing, natural language processing and sensor signal processing. In our case, the inputs are vectors, therefore, a CNN structure whose convolutional layers and max pooling layers are both one dimensional, is incorporated into our model.    

\subsection{Recurrent Neural Network Sub-structure}

Recurrent neural networks (RNNs) are widely used for analysing time-series issues, such as natural language process (NLP), computer vision and signal processing \cite{elman1990finding}. Unlike the hidden Markov model (HMM) \cite{krogh2001predicting}, RNNs can capture higher order dependency and has relatively less expensive computation. The state transition of a RNN can be expressed follow.  

\begin{equation}\label{Equ:RNN State}
h_t =\sigma_h(W_h*x_t+U_h*y_{t-1}+b_h)
\end{equation}

where, $h_t$ is the hidden state, $\sigma_h$ is the activation function, $W_h$ is the hidden weight, $x_t$ is the input, $U_h$ is the output weight, and $b_h$ is the bias. The output of a conventional RNN can be expressed as follow.   

\begin{equation}\label{Equ:RNN Output}
y_t =\sigma_y(W_y*h_t+b_y)
\end{equation}

where, $y_t$ is the output of RNN, $\sigma_y$ is the activation function, $W_y$ is the output weight and $b_y$ is the bias. 

Furthermore, a special variant of RNNs, the long short-term memory network (LSTM) \cite{gers1999learning} can solve the long-term dependency problem during learning process, which makes RNN even more powerful. More recently, the researchers proposed a type of LSTM, the gated recurrent unit (GRU) \cite{chung2014empirical}, which has almost the same accuracy as LSTM but less computing cost. In the following experiments, we will compare these three RNN structures as the sub-model of our approach.

The loss function is the mean square error (MSE) between RNN outputs and the training targets. Usually, such a MSE loss function is enough for many prediction problems. However, for our case, this type of loss function is not robust enough because the inputs and the outputs of our model have very complicated nonlinear relationship. 

\subsection{Mixture Density Network Sub-structure}

A traditional neural network with a distance-based loss function can be optimized by a gradient descent-based method. Generally, such scheme can perform quite well on the problems that can be described by a deterministic function $f(x)$, i.e., each input only corresponds to a output with one possible target value. However, for some stochastic problems like our case, one input may have more than one possible output values. Hence, this type of problems are better to be described as a conditional distribution $p(Y|X=x)$ than a deterministic function $Y = f(x)$. 

To tackle this issue, intuitively, we can replace the original distance-based loss function with a conditional probability function, for a regression task, the Gaussian distribution can be a appropriate choice. Moreover, using the mixed Gaussian distributions instead of a single Gaussian can improve the representation capacity of the model. Based on that, the researcher proposed the mixture density networks (MDNs) \cite{bishop1994mixture}. In contrast with traditional neural network, the output of MDNs is the parameters a set of mixed Gaussian distributions and the loss function become the conditional probabilities of given inputs. As a result, the optimization process is to minimize the negatived log probability. Hence, the loss function can be described as follow:

\begin{equation}\label{Equ: MDN Output}
P(y_t|x_t)=\sum_{k=1}^{K}{\pi_k P(y_t |x_t; \theta_k)}
\end{equation}

where, $\pi_k$ is the assignment portion for each sub-distribution, with $\sum_{k=1}^{K}\pi_k=1, (0<\pi_k<1)$, and $K$ is the total mixture models number. $\theta_k$ is the internal parameters of the base mixture distributions. For Gaussian distribution, $\theta_k =\{\mu_k,\sigma_k\}$, $\mu_k$ and $\sigma_k$ are the means and variances, respectively. Now, we can draw $y_t$ samples according to Eq. (\ref{Equ: MDN Output}) instead of computing $y_t$ directly based on Eq. (\ref{Equ:RNN Output}). In fact, Eq. (\ref{Equ:RNN Output}) will be used as the input of the MDN to depict the state transition. Thus, as the training process is finished, we can take use of the mixed Gaussian distributions to sample the target values according to the given inputs. To this end, we can use the maximum likelihood estimation (MLE), i.e, the means of the distributions are taken as the final prediction. In summation, the overall training process is depicted in Algorithm \ref{Alg: Algorithm}.     

\begin{algorithm} 
\caption{Algorithm}
\label{Alg: Algorithm}
\begin{algorithmic}[1]
\Require{$X$ (RSSI values)} 
\Ensure{$Y$ (coordinates)}

\While{e $<$ max epoch}
    \While{ i $<$ batch num}
    
    \State{$h_0$ $\gets$ $Conv1d(X)$} \Comment{convolutional operation}
    \State{$h_1$ $\gets$ max pool $h_0$}
    \State{$f$ $\gets$ flatten $h_1$} 

    \State{$h_{t}$ $\gets$ $\sigma_h(W_h*f_t+U_h*y_{t-1}+b_h)$} \Comment{update hidden states}
    \State{$\theta$ $\gets$ $\sigma_y(W_y*h_t+b_y)$} \Comment{compute network output}
    \State{$\theta_k$ $\gets$ split $\theta$} \Comment{assign mixture density parameters }
    \State{minimize the loss function: $-p(y_t|x_t;\theta)$}
   
    \EndWhile
\EndWhile\\

\State{$Y \sim p(y_t|x_t;\theta)$}\Comment{final output}
\Statex
\Return{$Y$}

\end{algorithmic}
\end{algorithm}

\section{Experiments and Results}
\label{sec: Experiments}

The implementation details of our model are illustrated in Table \ref{Tab: Implementation}. In the proposed model, the CNN sub-network consists a convolutional layer, a max-pooling layer and a flatten layer. The RNN sub-structure includes a hidden layer with $200$ neurons. The MDN sub-model is composed of a hidden layer and an output layer. The mixture model number for the MDN is $30$, and each mixture has $5$ parameters, the 2D means, variances and the mixture portions. As for the optimizer, according to \cite{arjovsky2017wasserstein}, for very nonstationary optimization problems, RMSProp \cite{tieleman2012lecture} can outperform Adam \cite{kingma2014adam}, thus we choose RMSProp as the optimizer.                

\begin{table*}[h]
\centering
\caption{Proposed model implementation details}
\label{Tab: Implementation}
\begin{tabular}{|c|c|c|c|}
\hline
Sub-network & Layer & Hyperparameter & Activation function\\
\hline

CNN & convolutional layer &  filter number: 100; stride: 2 & sigmoid\\
CNN & max pooling layer &  neuron number: 100  & relu \\
CNN & flatten layer &  neuron number: 100  & relu \\
RNN & hidden layer & memory length: 5; neuron number: 200 & sigmoid \\
MDN & hidden layer & neuron number: 200 & leaky relu  \\
MDN & output layer & 5*mixed Gaussians number (5*30) & - \\ 
\hline

\multicolumn{4}{|c|}{Optimizer: RMSProp; learning rate: 1e-3} \\

\hline
\end{tabular}
\end{table*}

In order to test our model on the WiFi fingerprints based sequential location prediction task, we conduct a series of experiments on the real-world dataset. We select two WiFi RSSI-coordinate paths from the Tampere dataset \cite{lohan2017crowdsourced}. This dataset includes a set of sequential RSSI value vectors with the input dimension of $489$. The detected RSSI values range from $-100$ dm to $0$ dm while the undetected WiFi access points are filled with value of $100$. Each vector has its own corresponding 2D coordinates labels. Therefor, for our task, the input is the RSSI values vector at current time point and the modeling target is the coordinates at next time point. 

\begin{figure}[!t]
\centering
\includegraphics[width= 1.\linewidth]{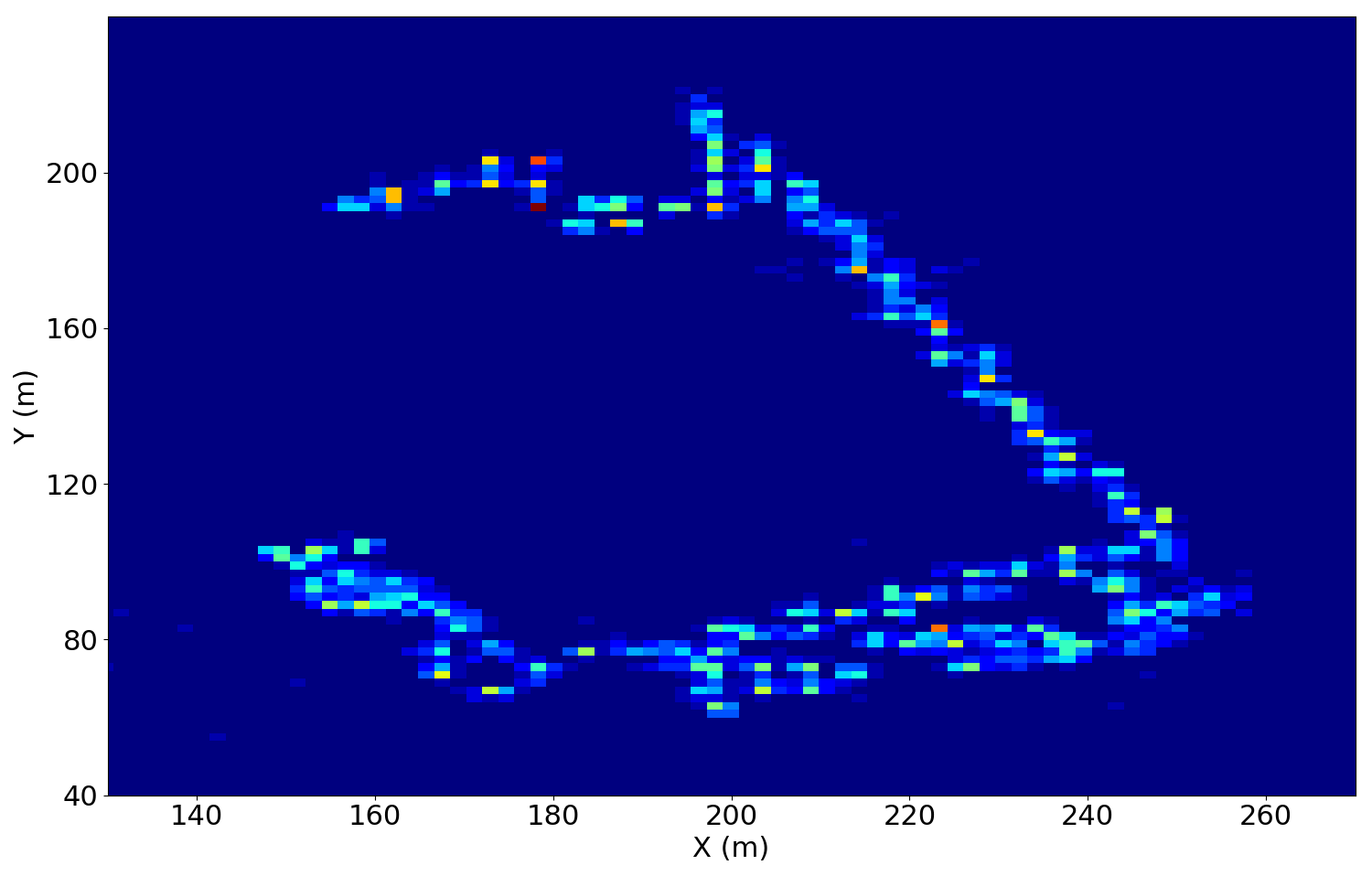}
\caption{Path 1 sampling prediction result by CMDRNN. }\label{Fig: Path_1}
\end{figure}

\begin{figure}[!t]
\centering
\includegraphics[width= 1.\linewidth]{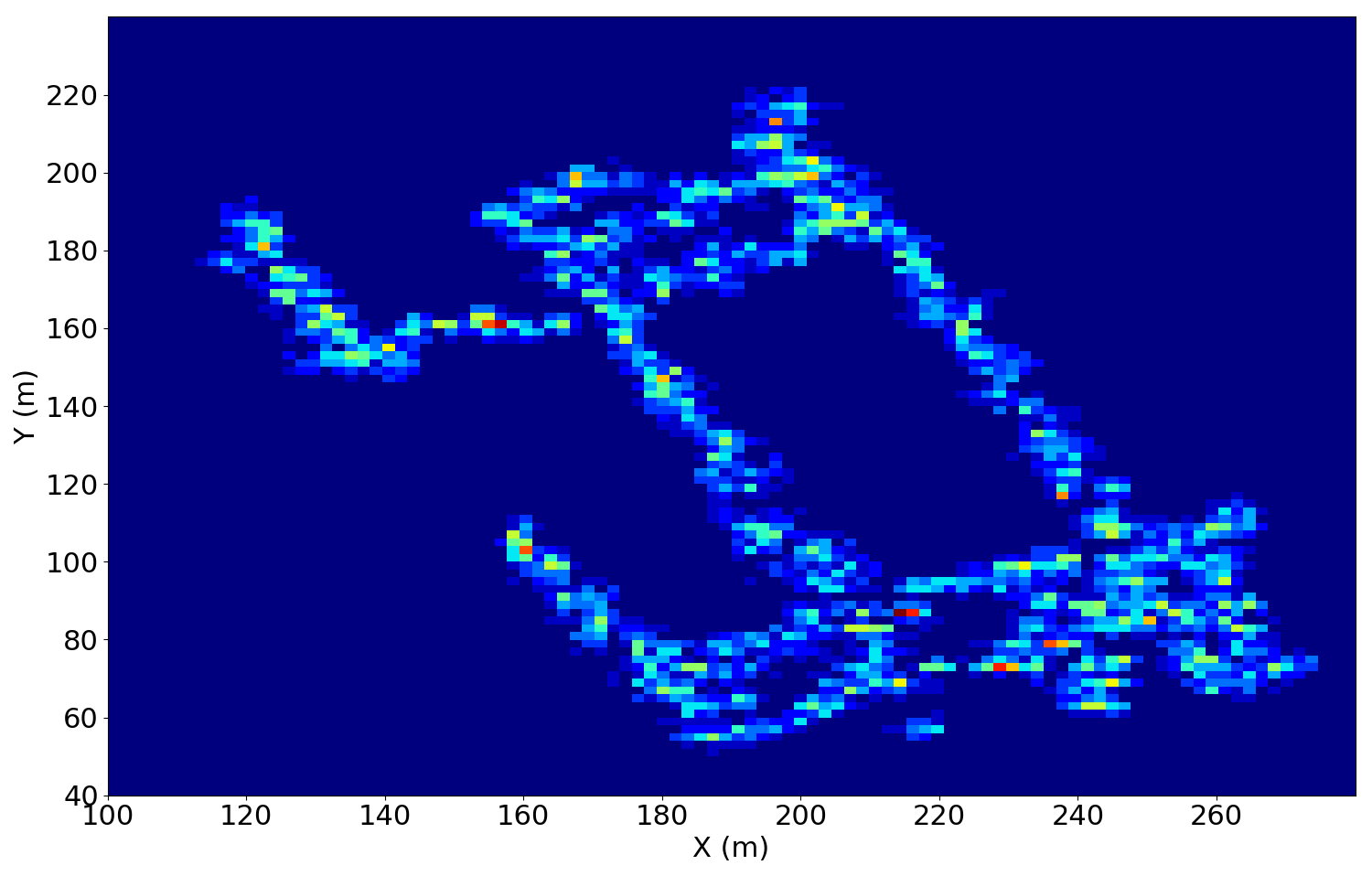}
\caption{Path 2 sampling prediction result by CMDRNN.}\label{Fig: Path_2}
\end{figure}

\begin{figure}[!t]
\centering
\includegraphics[width= 1.\linewidth]{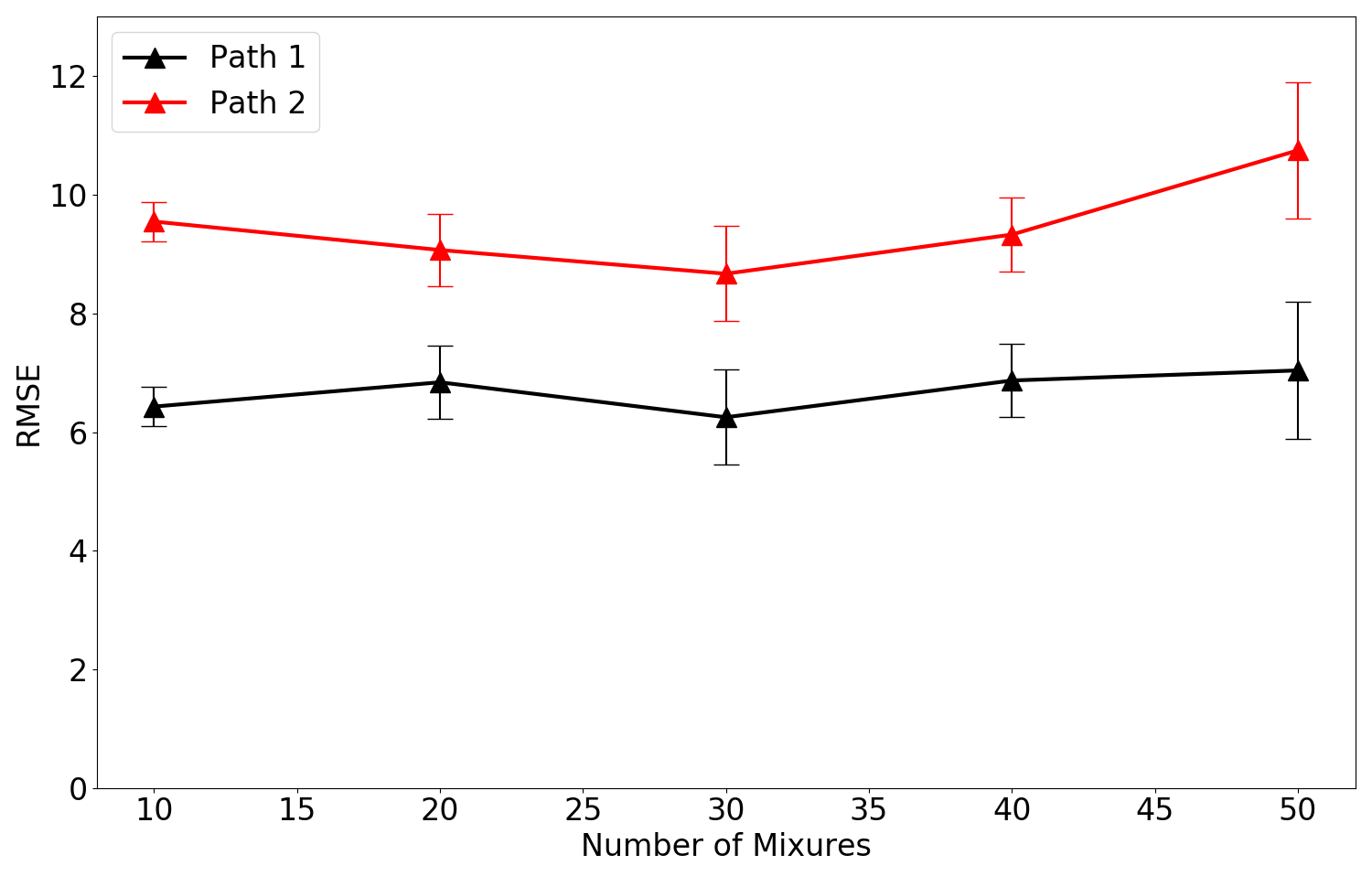}
\caption{Prediction results by varying mixture numbers in the MDN (bars represent the standard deviations).}\label{Fig: mixture_number}
\end{figure}

Fig. \ref{Fig: Path_1} and Fig. \ref{Fig: Path_2} show the prediction results of our proposed model. We also varies the mixture numbers in the MDN sub-model to find the optimal mixture number, which is $30$. The results are demonstrated in Fig. \ref{Fig: mixture_number}. Further, we compare our CMDRNN model to a set of deep learning approaches, RNN, CNN + RNN and RNN + MDN. The results are demonstrated in Table \ref{Tab: Path result}. From the experimental results, we can see that our proposed mode significantly improves the modeling accuracy compared to other deep learning methods on sequential user location prediction. Moreover, the GRU-based CMDRNN model has the best performance among the CMDRNN models.

\begin{table}[h]
\centering
\label{Tab: Path result}
\caption{Path prediction results (root mean square error)}
\begin{tabular}{|c|c|c|}
\hline
 Method & Path 1 & Path 2 \\
\hline
RNN     & $29.36 \pm 1.61 $ & $31.61 \pm 0.74$  \\
CNN+RNN & $34.26 \pm 3.04$ & $36.75 \pm 6.17$\\
RNN+MDN & $23.86\pm 5.50$ & $23.58\pm 2.29$ \\
CMDRNN(Vanilla-RNN)  & $8.26\pm 1.31$ & $10.17\pm 0.72$\\
CMDRNN(LSTM-RNN)  & $7.38\pm 0.89$ & $9.26\pm 0.31$\\
CMDRNN(GRU-RNN)  & $6.25\pm 0.80$ & $8.67\pm 0.23$\\

\hline
\end{tabular}
\end{table}

\section{Conclusions and perspectives}
\label{sec: Conclusions}

In this paper, we attempt to tackle the WiFi fingerprint-based user position prediction problem. In contrast with existing approaches, our solution is a novel hybrid deep-learning model. The proposed model is composed of three sub-deep neural networks, a CNN, a RNN and a MDN. This unique deep architecture takes advantage of the strengths of three deep learning models, which allows us to predict user location with high accuracy. For the validation, we tested our model on the real-world dataset, and the final results proves the effectiveness of our approach.

For the future work, we plan to exploit other deep generative models, for instance, variational autoencoders, Besysian neural network and normalising flows, for the potential applications on the WiFi fingerprint-based positioning problems. We should be aware that the labeled data is not always easy to acquire. As a matter of fact, in many cases, the available dataset are unlabeled. Hence, for the future work, we plan to investigate the semi-supervised learning techniques for human activity study.

\vfill\pagebreak



\bibliographystyle{IEEEtran}
\bibliography{refs}

\end{document}